# A Probabilistic Algorithm for Calculating Structure: Borrowing from Simulated Annealing

**Russ B. Altman, M.D., Ph.D.**
Section on Medical Informatics, MSOB X215
Stanford University Medical Center
Stanford, CA 94305
altman@camis.stanford.edu

## Abstract

We have developed a general Bayesian algorithm for determining the coordinates of points in a three-dimensional space. The algorithm takes as input a set of probabilistic constraints on the coordinates of the points, and an *a priori* distribution for each point location. The output is a maximum-likelihood estimate of the location of each point. We use the extended, iterated Kalman filter, and add a search heuristic for optimizing its solution under nonlinear conditions. This heuristic is based on the same principle as the simulated annealing heuristic for other optimization problems. Our method uses any probabilistic constraints that can be expressed as a function of the point coordinates (for example, distance, angles, dihedral angles, and planarity). It assumes that all constraints have Gaussian noise. In this paper, we describe the algorithm and show its performance on a set of synthetic data to illustrate its convergence properties, and its applicability to domains such as molecular structure determination.

## 1 MOLECULAR STRUCTURE

The determination of molecular structure is critical for many pursuits in biomedicine, including the study of how molecules perform their function and the design of drugs to interfere with this function [Stryer 1988]. As the human genome project produces large amounts of information about the atomic makeup of individual molecules, it becomes critical to devise methods for estimating molecular structure--that is, for determining how the atoms within molecules pack themselves in order to form three-dimensional structures.

The problem of determining molecular structure at a detailed level involves assigning three-dimensional coordinates to each of the atoms within a molecule. These molecules normally have on the order of 1000 to 10,000 atoms, and so we must estimate 3000 to 30,000 parameters to define a structure. The sources of information available for determining these structures are experimental, theoretical and empirical/statistical observations. They provide structural information ranging from geometric distances and angles to global measures of volume, shape and proximity to the surface. Typically, the problem of defining a structure is underdetermined for two reasons. First, there is insufficient information to position each atom uniquely. Second, the data sources are usually uncertain. It is therefore important to develop methods for combining evidence about structure that can handle the uncertainty explicitly. Moreover, it is critical that such methods produce not merely a single reasonable candidate structure, but also give some idea of the certainty associated with the position of each atom. There have been a few efforts to determine structure from a combination of experimental, statistical and theoretical data [Sippl 1990, Friedrichs 1991, Crippen 1990]. Not one of these methods is explicitly probabilistic. We have developed an algorithm, derived from the extended, iterated Kalman Filter, which can take a wide range of probabilistic constraints on structure and produce estimates of the mean and three-dimensional variance in the position of each atom. To avoid local optima, the algorithm uses a search heuristic—which is the same, in concept (although very different in implementation), as simulated annealing [van Laarhoven 1987, Vanderbilt 1984]. Simply stated, we iteratively estimate the positions of the points using all available data; we allow the algorithm to leave local optima by resetting the elements of a variance/covariance matrix elements to high values. By increasing the variance of the elements, we allow unsatisfied (relatively low-variance) constraints to make large changes in the estimates of location, and thereby to jump out of local optima. By iterating this process, we have been able to identify reliably sets of coordinates that satisfy the probabilistic constraints. We have already applied the algorithm to a number of experimental data-analysis tasks, and it has performed well[Pachter 1991, Pachter 1990, Arrowsmith 1991]. The goal of this paper is twofold: (1) to describe the methodology in detail, while explaining how we have borrowed concepts from simulated annealing to improve the convergence properties; and (2) to investigate the sensitivity of the method to perturbations in the reheating heuristic. The ideas used in our work should be useful in a variety of settings where probabilistic algorithms are searching a large space.

# 2  THE DATA REPRESENTATION

There are three types of information that our algorithm uses: an estimate of the mean position of each point (or atom), an estimate of the variance/covariance between all coordinates of all points, and a representation of the underlying model of the data and its noise. The notation used here is borrowed from Gelb [Gelb 1984] and from Smith et al [Smith 1986].

For molecular structure, the parameters to be estimated are the coordinates of atoms in three-dimensional space. We represent the mean positions of each atom as a vector, $\mathbf{x}$, of length $3N$ for $N$ atoms:

$$\mathbf{x} = [\, x_1 \ \ y_1 \ \ z_1 \ \ x_2 \ \ y_2 \ \ z_2 \ \ldots \ x_N \ \ y_N \ \ z_N \,]^T \qquad [1]$$

The second element of our representation is a variance/covariance matrix for vector $\mathbf{x}$. This matrix, $\mathbf{C(x)}$, contains the autocovariance information for vector $\mathbf{x}$: the diagonal elements contain the variances of each element of $\mathbf{x}$, whereas the off-diagonals contain the covariances between the elements within $\mathbf{x}$:

$$\mathbf{C(x)} = \begin{pmatrix} \sigma^2_{x_1} & \sigma^2_{x_1 y_1} & \cdot & \cdot & \sigma^2_{x_1 z_N} \\ \cdot & \sigma^2_{y_1} & & & \cdot \\ \cdot & & \cdot & & \cdot \\ \cdot & & & \cdot & \cdot \\ \sigma^2_{z_N x_1} & \cdot & \cdot & \cdot & \sigma^2_{z_N} \end{pmatrix} \qquad [2]$$

Because the coordinates can be logically grouped into triplets (representing the x,y,z coordinates for a single atom), we abbreviate our notation of $\mathbf{C(x)}$ as a matrix with submatrices.

$$\mathbf{C(x)} = \begin{pmatrix} C(x_1) & \cdot & \cdot & C(x_1 x_N) \\ \cdot & C(x_2) & & \cdot \\ \cdot & & \cdot & \cdot \\ C(x_N x_1) & \cdot & \cdot & C(x_N) \end{pmatrix} \qquad [3]$$

where each of the submatrices represents the variance of a single atom, or the covariance between two atoms.

$$C(x_i x_j) = \begin{pmatrix} \sigma^2_{x_i x_j} & \sigma^2_{x_i y_j} & \sigma^2_{x_i z_j} \\ \sigma^2_{y_i x_j} & \sigma^2_{y_i y_j} & \sigma^2_{y_i z_j} \\ \sigma^2_{z_i x_j} & \sigma^2_{z_i y_j} & \sigma^2_{z_i z_j} \end{pmatrix} \qquad [4]$$

The utility of our representation for structural display can now be understood: the mean values for the coordinates of each atom can be taken from the vector, $\mathbf{x}$, and plotted. In addition, the variance of each coordinate of an atom can be extracted from the diagonal and provides the uncertainty along each axis of the mean estimate. In fact, with the full *3 x 3* variance/covariance information, we can estimate the uncertainty in any direction. Figure 5 shows the molecule with mean positions attached to form a backbone trace, and selected uncertainties displayed to show two standard deviations of variance.

The off-diagonal elements of the variance/covariance matrix contain information about the dependence between the coordinates of two atoms (that is, the dependence of the position of one atom on the position of the other). Each off-diagonal element is a linear estimate of the relationship between two coordinates. It is related to a correlation coefficient by a normalization term. If the element is positive, then the two coordinates are positively correlated. This information is critical to the search: a change in any atom position affects the position of other atoms through this first order estimate of their covariation. Thus, the off-diagonal *3 x 3* submatrices represent a summary of how the position of one atom changes as the position of another is modified. There is a strong network aspect to this representation. As more is learned about the relationships between atoms, the network of dependencies grows (for example, see Figure 4). Eventually, the movement of any atom results in the concerted movement of all other atoms based on this covariance information. The precise mechanisms for updating estimates of the mean vector and covariance matrix are discussed in the next section.

## 2.1  REPRESENTATION OF CONSTRAINTS

We take a constraint to be any information that constrains the possible values of the coordinates. In general, we model constraints in the following form:

$$\mathbf{z} = \mathbf{h(x)} + \mathbf{v} \qquad [5]$$

where $\mathbf{z}$ is the measured constraint (that is, the value provided by the experimental, theoretical or statistical source of information). $\mathbf{z}$ can be scalar or vector. It is modeled as having two parts: the first part is an (in general) vector function, $\mathbf{h(x)}$, which is a function of the mean vector, $\mathbf{x}$. The second part of the model, $\mathbf{v}$, models the noise in the system. Given a perfect measurements, $\mathbf{v}$ is zero and the measured constraint takes on the exact value of the model function, $\mathbf{h(x)}$. In general, $\mathbf{v}$, is a Gaussian noise term which models the degree of certainty in any given measurement.

Thus, for example, a measurement of distance between two points would be represented as a function of 6 elements of the mean vector, $\mathbf{x}$:

$$z = (( (x_i - x_j)^2 + (y_i - y_j)^2 + (z_i - z_j)^2)^{0.5} + \mathbf{v} \qquad [6]$$

If the distance measurement refers to the distance between two carbon atoms in a covalent bond, then the variation in $\mathbf{v}$ is extremely small (the covalent bond distance varies less than 0.1 Ångstroms). If the distance measurement refers to an experimental measurement from, for example, a study using Nuclear Magnetic Resonance (NMR), then $\mathbf{v}$ will have larger variation (NMR distances vary as much as 5 Å) [Wuthrich 1985]. For the purposes of this paper, we shall use distance measurements with different variances to illustrate the performance of the algorithm. We have shown elsewhere [Arrowsmith 1991, Liu 1992] that the model is general and extends to bond angles (a nonlinear function of nine coordinates), and dihedral angles (a nonlinear function of twelve coordinates).

## 2.2 INTRODUCING CONSTRAINTS

Having established our representation for atomic position, atomic uncertainty, and constraints, we can understand the mechanism for introducing constraints and updating our estimates of the state vector, $\mathbf{x}$, and the covariance matrix $\mathbf{C}(\mathbf{x})$. A standard Kalman filter employs an update algorithm of the following form [Gelb 1984]:

$$\mathbf{x}(new) = \mathbf{x}(old) + \mathbf{K} \{ \mathbf{z} - \mathbf{h}(\mathbf{x}(old)) \} \qquad [7]$$

where

$$\mathbf{C}(new) = \mathbf{C}(old) - \mathbf{K} \, \mathbf{H} \, \mathbf{C}(old) \qquad [8]$$

$$\mathbf{K} = \mathbf{C}(old) \, \mathbf{H}^T \{ \mathbf{H} \, \mathbf{C}(old) \, \mathbf{H}^T + \mathbf{v} \}^{-1} \qquad [9]$$

and

$$\mathbf{H} = \frac{\delta \mathbf{h}(\mathbf{x})}{\delta \mathbf{x}} \Big|_x \qquad [10]$$

(In general, the Kalman filter allows for a time-dependent modeling of how $\mathbf{x}$ and $\mathbf{h}(\mathbf{x})$ change. We assume a static molecule and do not introduce any time-dependent model of change. We therefore are interested in calculating a single estimate that, for example, corresponds to time = 0.)

Simply stated, the new estimate of mean position is the weighted difference between the old estimate of mean position and the new positions implied by the new data. If the measured data, $\mathbf{z}$, have high variance compared with our estimated variance in our certainty in $\mathbf{h}(\mathbf{x})$, then the new state estimate will not be updated by much. If, on the other hand, the measured data, $\mathbf{z}$, have low variance compared with the estimate, $\mathbf{h}(\mathbf{x})$, produced by the current coordinate estimates, then the new coordinate estimate will be substantially different from the current estimate.

In general, the magnitude of the update in the coordinate vector will reflect the relative certainties of the measured and predicted values for each constraint. Thus, early in the problem solving, when few constraints have been introduced (assuming serial introduction of constraints), it is relatively easy to move atoms around because they have a large initial covariance. Later in problem solving, however, when the estimate of the uncertainty in the positions is lower, it is much more difficult to move atoms unless there are very certain (that is, low-variance) measurements.

The Kalman filter, as discussed earlier, is an optimal estimator when $\mathbf{h}(\mathbf{x})$ is linear [Gelb 1984]. When $\mathbf{h}(\mathbf{x})$ is nonlinear, the extended, iterated Kalman filter can be used. This version of the filter performs a local search iteratively around $\mathbf{h}(\mathbf{x})$ to find the locally optimal value for $\mathbf{x}$:

$$\mathbf{x}(new)_i = \qquad [11]$$
$$\mathbf{x}(old) + \mathbf{K}_i \{ \mathbf{z} - \mathbf{h}(\mathbf{x}(new)_{i-1}) + \mathbf{H}(\mathbf{x}(old) - \mathbf{x}(new)_{i-1}) \}$$

(where $i$ is the local iteration counter). $\mathbf{C}(\mathbf{x})$ is as defined in Equation 8.

The extended, iterated Kalman filter has been shown to be the optimal linear approximation to the actually nonlinear solution. Unfortunately, the nonlinearities of structural determination are such that the residual errors of structures calculated by sequentially introducing constraints are still too large. However, the solution produced by the extended, iterated Kalman filter is typically satsifies the input constraints better than the starting estimate (but not good enough). Initially, we had hoped that, by serially introducing the constraints a second time, we would continue to improve our estimate. However, because the covariance estimates decrease monotonically with the introduction of each constraint, it becomes harder and harder to move atoms out of the local minima defined by the extended, iterated Kalman filter.

## 2.3 HEATING TO AVOID LOCAL OPTIMA

Our approach to the solution of the problem of local minima was inspired by the work in simulated annealing [van Laarhoven 1987, Vanderbilt 1984]. After serially introducing the constraints on the structure, we are left with an improved estimate of $\mathbf{x}$, but also a covariance matrix $\mathbf{C}(\mathbf{x})$ that is unwilling to allow atoms to move out of the local minima, in the sense that extremely low-variance measurements would be required to move an atom far from the estimate. We introduce "heat" by resetting the covariances to their initial (large) values, and thus allow unsatisfied constraints to have a relatively greater effect on the vector, $\mathbf{x}$. We then reintroduce all the constraints once again, but sorted such that the constraints that were least satisfied by the previous coordinate estimates are introduced into the solution earliest. We believe (and have tested in experiments described later in this section) that by reheating the covariances and introducing the constraints in reverse order of satisfaction, we maximize the chance that the atoms will be reorganized radically and will jump out of the current minima. Since we have observed a consistent ability of the iterated, extended Kalman filter to improve upon its starting estimates, we simply repeat the cycle of search, reheat, search until the residual errors are acceptable (see outline of procedure). Although we have not made any formal claims about resistance to minima, there are three forces acting to help the algorithm avoid (or leave) multiple minima. First, we use a covariance matrix to capture the first-order correlation between atomic coordinates; therefore, moving even a few atoms causes changes in the entire molecule (and more coverage of the search space). Second, the reheating allows atoms to move from one local minima to another in a rational way: Atoms will not move arbitrarily in space, but will move along a vector whose magnitude is consistent with the known initial excursion of the atom. Third, the reordering of serial constraints allows the constraints that are violated to dominate the initial reorganization of the structure. The performance of our algorithm is illustrated in the next section with an example.

```
Procedure for Double Iterated Kalman Filter

0.    Initialize X = Xo
         (e.g., Xo = random coordinates if no
         other information available)

      Initialize C(X) = C(Xo)
         (e.g., C(Xo) typically a diagonal,
         uncorrelated matrix with elements
         compatible with the overall anticipated
         size of molecule)

1.    Cycle Number = 0
2.    Increment Cycle Number
3.    For each constraint, Z(i), i = 1,N,

         Update estimate of X and C(X) using
         extended, iterated Kalman Filter for
         constraint Z(i) yielding X' and C(X')

4.    Evaluate average error <E(i)>,
      and maximum error Max[E(i)], where

         E(i) =  Z(i) - H(X')
                 ───────────
                  Sqrt(v(i))

5.    Check end conditions on average, maximum
         error.  End if done:

         Solution = X', C(X')

6.    If not done,
         set X = X',
            C(X) = C(Xo)    ;REHEAT STEP
7.    Sort constraints
8.    Goto Step 2.
```

In this paper, we illustrate the convergence properties of the method under conditions of noise and sparse data. We also investigate the importance of the sorting step (Step 7) so that we can better understand the mechanism whereby the reheating step improves convergence. We are testing the hypothesis that, by sorting the constraints in reverse order of satisfaction, we increase the ability of false optima solutions to be abandoned. Alternatively, the heating itself may be the only key step, and the order of constraints may not be critical.

## 3  EXPERIMENTS PERFORMED

To illustrate the performance of the program on a typical data set, we have chosen the problem of defining the topology of a small protein, crambin [Hendrickson 1981]. Crambin contains roughly 500 atoms, but for the purposes of this example, we considered only the 46 backbone heavy atoms that define the general topology of the molecule. The structure of crambin is known, so we generated synthetic data sets for these tests. In general, there are 1035 distances between 46 atoms. The minimum number of exact distances required to define the position of $N$ points is $4N-10$, or 526 is the case of crambin. The state (coordinate) vector, therefore, has 134 parameters and the covariance matrix is 134 $x$ 134. For all calculations, the starting values for the x vector were generated randomly between 0 and 50 Ångstroms. The covariance matrix was initialized to have all diagonal elements at 100 (that is, a starting variance of 100 $Å^2$ for each atom), and off-diagonal elements set to 0 (implying independence of all coordinates initially)[1]. We could introduce a less random initial model, if we had information about the type of structure we might be expecting. In this case, we assume no critical prior information about the structure. We performed the following tests:

Test 1:  To test the maximal speed of convergence, we tested the algorithm by providing all possible exact distances (1035), with extremely low variance.

Test 2:  To explore the dependence of the reheating step on the order of constraints, we provided 33% of possible distances (334) , with very low variance and employed the following strategies for processing constraints:
   a: sorted constraints in decreasing order of error
   b: randomly shuffled constraints
   c: left constraints in constant, fixed order

Test 3:  To explore the effects of noise on the system, we provided the same 33% of possible distances, but added noise.  We sorted all constraints as in test 2a.

---

[1] As constraints between atoms are introduced and propagated, they gain non-zero covariances.

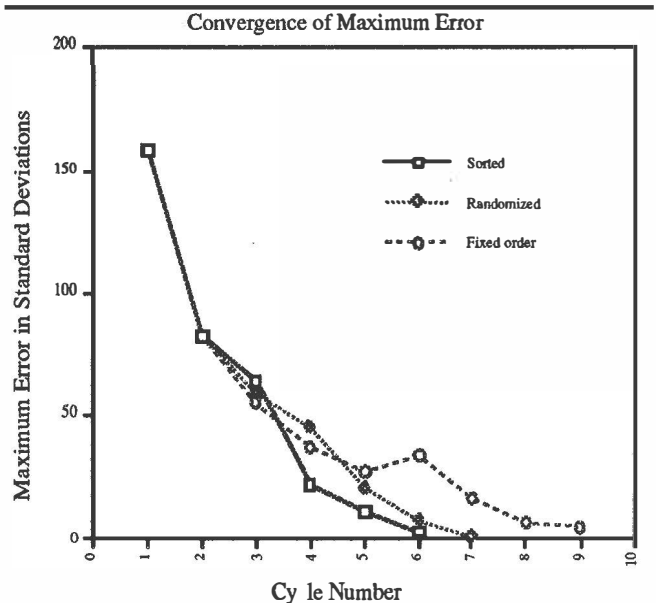

Figure 1. Each of three strategies for ordering constraints is compared with respect to the maximum error of all constraints as a function of Cycle number.     Sorted constraints were introduced in reverse order of satisfaction at Step 7 of the algorithm as defined in the paper. Randomized constraints were introduced in random order. Fixed order constraints were introduced in the same order for each cycle.

a:  We introduced low noise into the measurements. For each constraint a variance was randomly selected between 0.0 and 6.0 $Å^2$, and then a noise term was selected from a normal distribution with this variance, and was added to the measurement.

b:  We introduced high noise into measurements as in Test 3A, but variances ranging from 0.0 to 25.0 $Å^2$.

Test 4:  To explore the effects of noise and sparse data, we provided a set of 10% of possible distances chosen at random.

a:  We provided exact distances
b:  We added an average of 3.0 Å to each distance

For all runs, the tolerance for exiting the inner loop of the iterated, extended Kalman filter was 0.01, and the maximum number of cycles, $i$, was three. The stopping condition for all runs (unless otherwise noted) was an average error of 0.3 SD or a maximum error of 1.0 SD.

# 4   RESULTS

Test 1:  The starting structure had an average error (in SD from measured value) of 60, with a maximum error of 175.  With all possible exact distances, the algorithm converged to an average error of 0.20 SD (maximum error 1.3 SD) in 3 cycles.  To test the stability of the solution, we allowed the algorithm to run for a total of 1000 cycles. The solution remained stable, and the ultimate improvement to an average error of 0.0007 SD (maximum of 0.002) was achieved at cycle 58.  Cycles 59 through 1000 made no further improvement.  The structure that resulted was identical to the target solution, as expected.

Test2:  With 33% of all exact distances, the starting average error was 62 SD (maximum, 158 SD).  Figures 1 and 2 illustrate the performance of each of the three strategies for ordering constraints.    Sorting in reverse order of error (in SDs) consistently lead to the quickest convergence.    A fixed order of constraints consistently converged poorly.   The peak at cycle 6 for the maximum error of the fixed order constraints represents a jump out of a local minima at cycle 5, with subsequent convergence by cycle 9.     All three methods produced structures that matched the solution to a root mean squared distance (RMSD) of 0.01 Å.

Test 3:  The starting average error for all tests 3a and 3b were (on average) 16 SD and 9 SD, and (maximum) 224 and 360 SD.  Because of the noise, no structures were found that satisfied the end conditions.  However, test 3a yielded a best solution with average error 0.58 SD (maximum: 2.53 SD).  Test 3B yielded a best solution with average error 0.76 SD (maximum: 2.81 SD).  More important, the structures generated by these data sets matched the target solution to a RMSD of  2.0 Å and 3.7 Å, respectively.  Since the maximum dimension of the molecule is 30 Å, these represent good estimates of structure, and clearly show the general topology of the molecule. Figure 4 shows a contour diagram of the final covariance matrix for test 2a. It provides a graphical diagram of the dependencies contained within the data. It is clear, for example, that there is a strong correlation between the position of atoms 5 and 45, and along all the sequential neighbors along the diagonal. There are also regions of correlation stretching from atoms 6 to 26 and from atoms 36 to 44. There is a clear lack of correlation between the region around atom 19 and that around atom 40, and to a lesser extent in the regions surrounding atoms 5 and 20.

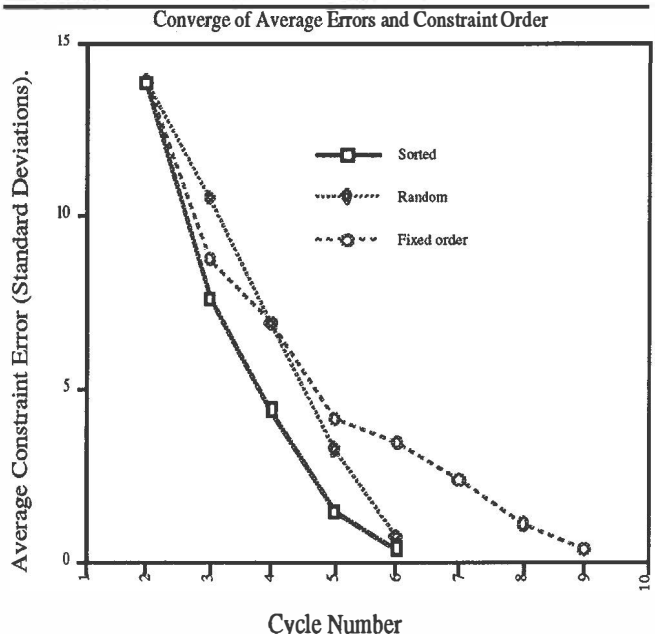

Figure 2. Each of three strategies for ordering constraints is compared with respect to the average error of all constraints as a function of Cycle number.    Sorting strategies were as described in Figure 1

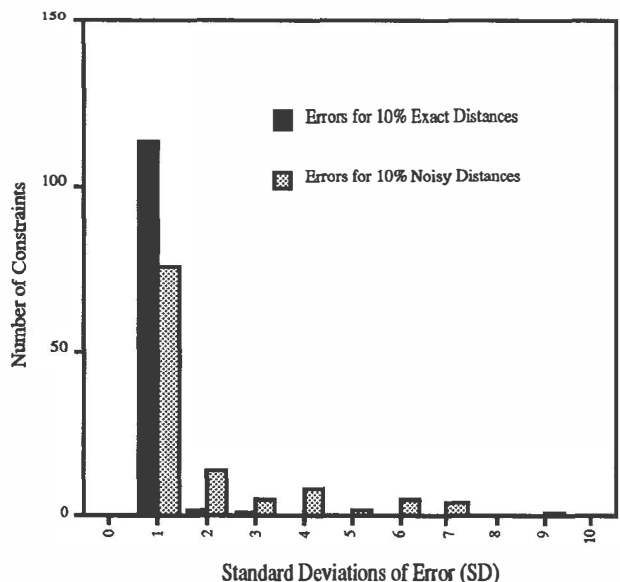

Figure 3. The errors for constraints used in Tests 4A and 4B plotted as histogram. Error is defined as $(z(i) - h(x))/\sqrt{v(i)}$. The 10% data set with exact distances has very low residual errors, whereas the noisy data set has a few constraints with errors above 4 SD.

Test 4: In the case of 10% of all possible exact distances, the initial errors were 71 SD (average) and 158 SD (maximum). The algorithm converged to an average error of 0.36 SD with a maximum of 1.9 SD. The structure matched the target solution to an RMSD of 2.13 Å. In the case of the same 10% of distances with noise added, the initial errors were 36 SD (average) and 151 SD (maximum). The average error was 1.2 SD, with a maximum of 8.5 SD. The structure resulting from the noisy data has an RMSD of 5.13 Å. Figure 3 shows the distribution of errors for each of these two calculations.

## 5  DISCUSSION

The Kalman filter has proven to be a versatile tool since its inception in the early 1960s. It is an optimal filter for linear problems. For nonlinear problems, it has been shown to be the optimal linear solution [Gelb 1984]. The extended, iterated Kalman filter has proven useful as a suboptimal estimator for nonlinear problems as well. However, it is suboptimal, and for problems requiring high accuracy, it can fall short. In this paper, we have shown that resetting the covariance matrix allows the algorithm to converge reliably in nonlinear problems such as molecular structure based on distance information. Whereas simulated annealing protocols require a cooling procedure to induce equilibration, our algorithm allows the serial introduction of constraints to cool the structure. We have not yet experimented with more sophisticated ways to reheat the covariance matrix. However, the current method simply throws away the covariance information derived during a single cycle (cf. step 6 in the procedure outline), and we are investigating more intelligent ways to reheat. We have shown that the performance degrades gracefully as noise is added, and constraints are subtracted. Both with 33% and 10% of all distances (and with noise added), we were able to find reasonable topologies for the test molecule.

Our experiments with different constraint orders confirm our hypothesis that the reheating of the covariance matrix allows the solution space to be explored more effectively. A fixed order of constraints is more likely to explore the same general hypothesis space, and to converge more slowly than either a random order or an order in which the most dissatisfied constraints take the "first shot" at altering the solution.

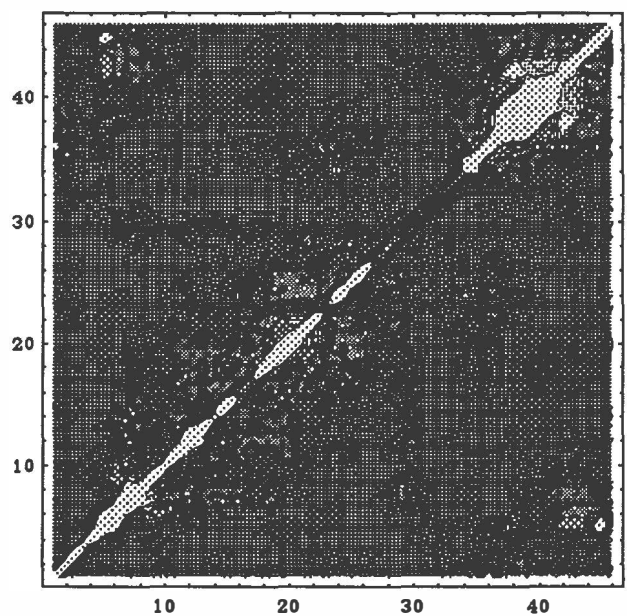

Figure 4. A contour representation of the covariance matrix between the 46 atoms that were placed in Test 3A. Each value is the norm of the submatrix $C(x_{ij})$ of the variance/covariance matrix $C(x)$. The variances of the individual atoms (along the diagonal) are for the most part, larger than the covariances between atoms. There is a region of particularly high covariation between atoms 45 and 5. There is a region of low covariation between atoms 40 and 18.

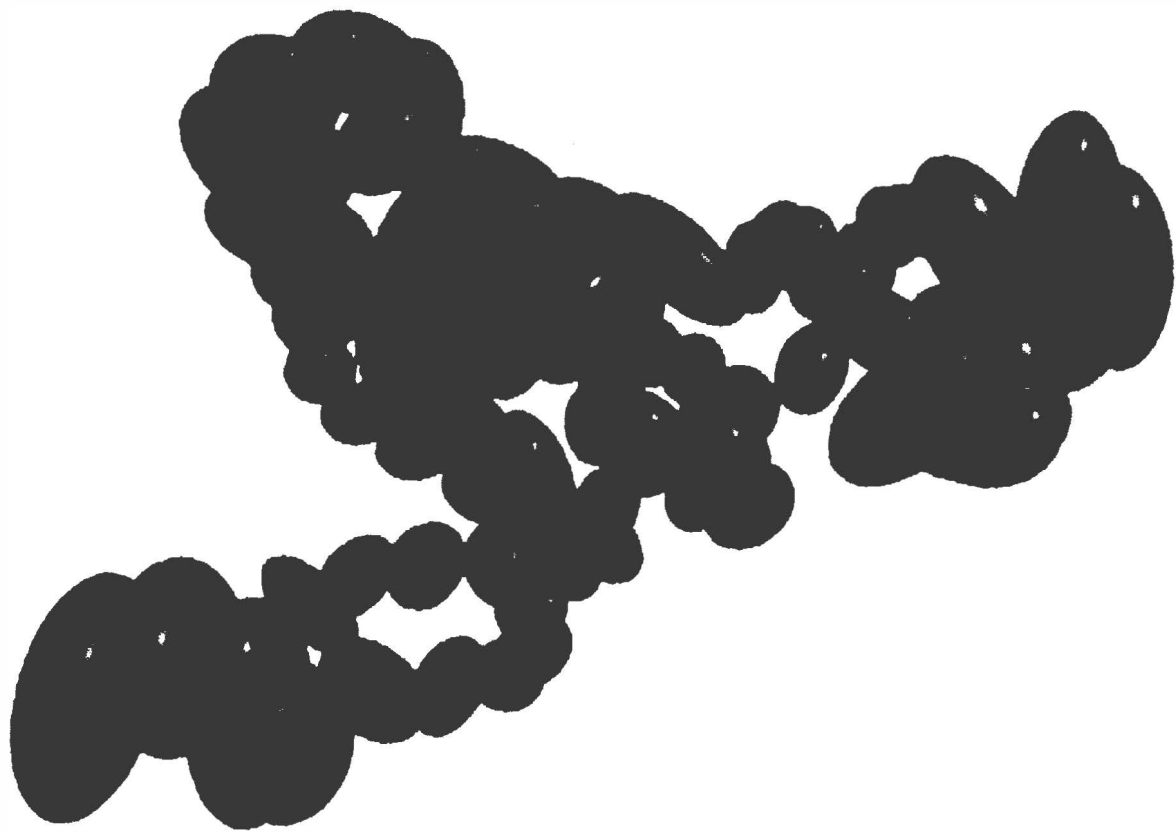

Figure 5. Sample structure produced by the algorithm. The atoms are connected with thin lines, and an ellipsoid of uncertainty (drawn at 2 SD assuming three-dimensional Gaussian distribution) is displayed for a subset of atoms. The variance in the position of atoms clearly is variable: well-defined atoms (such as in lower left) have smaller ellipsoids than poorly constrained atoms (such at far right).

One advantage of our method is that the ending conditions for the search (average or maximum SD or errors) are not part of the machinery of search. We can therefore easily change our error metric and end conditions (step 5 in the procedural outline). Usually, we search for the x vector that has the minimum average error. However, the error profile of the resulting structure (such as shown in Figure 4) is actually better than the variances on the constraints imply! That is, our distribution of errors should reflect the variance of the measurements in these errors, but they do not. We may, therefore, be overfitting our data. In the case of extremely precise measurements (such as test 1, 2a,bc, and 4a) overfitting is difficult to demonstrate because the answer is so precisely constrained. When we have extremely sparse or noisy data, however, we may discard superior solutions. Current work focuses on using other metrics to evaluate possible solutions.

Although we have concentrated in this paper on the use of distance constraints between points, the mathematical form of the filter makes it clear that (1) any function of the coordinates can be used as a constraint model, and (2) these functions need not be scalar, but rather can be vector functions. In our work so far, we have limited ourselves to distances, angles, and dihedral angles because these types of constraints are sufficient for most structure determinations from NMR data. However, as we collect statistical data on the associations between certain types of atoms and aggregates of atoms, we can use statistical distributions as constraints on our molecule. With respect to the use of vector functions, we have avoided them so far (and preferred the serial introduction of scalar constraints) because this allows the inverse term in equation 9 to remain a scalar, and thus avoids a matrix inversion--which would be computationally expensive (to do or to avoid). We have run tests on the serial introduction of constraints versus parallel introduction (for example, the vector z can contain all the distance measurements, and v contains all the variances). We have found that, with all the information available, a greater improvement in the solution occurs per cycle, but that the cost per cycle increases such that the two strategies are roughly equivalent.

## 5.1 RELATED WORK

There are two lines of research that are related to the work described here. The first is that of molecular structure determination. Distance geometry, is an algorithm which takes as input a set of distances between atoms within a molecule. It employs a clever eigenanalysis of a matrix derived from these distances to estimate the coordinates of the structure [Havel 1984, Havel 1983]. It only refers to min/max boundaries on parameter values for input, and produces as output a single solution. To estimate the uncertainty in the structure, it is necessary to run the algorithm many times and collect statistics over the population of structures. Energy minimization and molecular dynamics are algorithms for structure determination which are based on the assumption that the proper conformation of a molecule is the one that has the lowest free energy [Nemethy 1990, Levitt 1988]. Energy terms that describe the interactions between all pairs of atoms within a structure can be defined, and optimization methods can be applied to find the conformation of the structure that has the lowest energy. Uncertainty is represented to some extent within the energy profiles (which are related to probabilities by the Boltzmann relation). These algorithms are based on physical forces, and it is difficult to know how to consider probabilistic sources of data. A detailed comparison between our algorithm and these other approaches (varying constraint abundance, precision of constraints, size of molecule) has been published and shows that the time for a single DIKF calculation to converge is equivalent to roughly 10 molecular dynamics runs and 100 distance geometry runs [Liu 1992].

The second line of research related to our work is that in optimization and parameter estimation [Gelb 1984]. The relationship of our method to the standard extended, iterated Kalman filter is clear: We are simply re-iterating and reintroducing constraints after adjusting the covariance matrix and reordering our constraints. Our method therefore is a member of the class of nonlinear least-squares–like estimator that seeks the most likely set of coordinates that best satisfy the input constraints. It is Bayesian because it uses an initial model of the solution (in these experiments the model was randomly generated). Our method uses a first-order approximation to the nonlinearities of the system, and improves its performance by iteration. Simulated annealing is a computational method for assisting optimization by providing a powerful heuristic for efficient search [van Laarhoven 1987, Vanderbilt 1984]. Based on an analogy to solid-state physics, simulated annealing protocols add "heat" to an optimization to increase the likelihood that a solution will jump out of a local optima. The solution is then allowed to "cool" slowly such that it settles into a new optimum—as a cooled solid might settle into a new crystalline packing. Our method shares many high level concepts with simulated annealing: By increasing the variances and covariances, we are increasing the range of possible values for each parameter, and by introducing the constraints in reverse order of satisfaction, we give the least satisfied constraints a chance to pull the solution out of a local minima. Although it is heuristic in nature, we have found that this protocol reliably finds low average error structures, as well as low maximum errors [Pachter 1990, Liu 1992].

## 6    CONCLUSIONS

The chief limitations to our method are the assumption of Gaussian noise in the constraints, and the assumption of Gaussian distribution when drawing the calculated structures. Gaussian noise is clearly not a valid assumption for many constraints that we would like to use. There is a body of literature on the use of Kalman filters with non-Gaussian noise, and we are currently investigating its applicability to our technique [Sorenson 1970]. Our algorithm produces a two-moment estimate of atomic location (three-dimensional mean and variance). For purposes of display, we assume that these represent the first two moments of a three-dimensional Gaussian when drawing atomic locations. Of course, it is possible that some atoms will have a bimodal distribution, and we are unable to capture these distributions currently. Moving to multimodal representations of atomic position is not a priority in our work for two reasons. First, as more independent data sources are introduced, the three-dimensional Gaussian becomes the most likely final distribution by the central limit theorum. Second, there are few biological examples of significant bimodal distributions.

In this paper, we have introduced our algorithm, illustrated its performance on a set of test problems, and investigated convergence speed as a function of the order in which constraints are introduced. We have shown that:

1. A probabilistic formulation of molecular structure determination is natural, and provides a common language in which multiple data sources can be combined.

2. The double iterated kalman filter (DIKF) represents one possible implementation of an engine to determine probabilistic structures. It employs a search heuristic similar to simulated annealing, and is able to converge to reasonable solutions under a variety of circumstances, including data with varying levels of noise and abundance.

3. The DIKF's mechanism for sorting constraints in order of decreasing error (after reheating) leads to a more efficient search for low average error structures, when compared with a random and fixed order of constraints. Each of the three strategies produces the same result eventually, but with a different efficiency. These results support the contention that the reheating step works because it allows local optima to be avoided.

## Acknowledgements

This work was supported by computing resources provided by the Stanford University CAMIS project, which is funded under grant number LM05305 from the National Library of Medicine of the National Institutes of Health. RBA also received a hardware grant from Hewlett-Packard. Graphical display software provided by Steve Ludtke (Ludtke 1993).